\def\year{2021}\relax

\documentclass[letterpaper]{article} 
\usepackage{aaai21}  
\usepackage{times}  
\usepackage{helvet} 
\usepackage{courier}  
\usepackage[hyphens]{url}  
\usepackage{graphicx} 
\usepackage{amsmath,amsfonts,mathtools}
\usepackage{algorithm}
\usepackage{algorithmic} 
\usepackage{array}
\usepackage{arydshln}
\usepackage{graphicx}
\usepackage{booktabs}
\usepackage{xcolor}
\usepackage{multirow}
\usepackage{rotating,tabularx}
\usepackage{svg}
\usepackage[switch]{lineno} 

\makeatletter

\newcommand*{\Xbar}{}%
\DeclareRobustCommand*{\Xbar}{%
  \mathpalette\@Xbar{}%
}
\newcommand*{\@Xbar}[2]{%
  \sbox0{$#1\mathrm{X}\m@th$}%
  \sbox2{$#1X\m@th$}%
  \rlap{%
    \hbox to\wd2{%
      \hfill
      $\overline{%
        \vrule width 0pt height\ht0 %
        \kern\wd0 %
      }$%
    }%
  }%
  \copy2 %
}
\newcommand*{\gbar}{}%
\DeclareRobustCommand*{\gbar}{%
  \mathpalette\@gbar{}%
}
\newcommand*{\@gbar}[2]{%
  \sbox0{$#1\mathrm{g}\m@th$}%
  \sbox2{$#1g\m@th$}%
  \rlap{%
    \hbox to\wd2{%
      \hfill
      $\overline{%
        \vrule width 0pt height\ht0 %
        \kern\wd0 %
      }$%
    }%
  }%
  \copy2 %
}
\newcommand*{\Hbar}{}%
\DeclareRobustCommand*{\Hbar}{%
  \mathpalette\@Hbar{}%
}
\newcommand*{\@Hbar}[2]{%
  \sbox0{$#1\mathrm{H}\m@th$}%
  \sbox2{$#1H\m@th$}%
  \rlap{%
    \hbox to\wd2{%
      \hfill
      $\overline{%
        \vrule width 0pt height\ht0 %
        \kern\wd0 %
      }$%
    }%
  }%
  \copy2 %
}

\usepackage{natbib}  
\usepackage{caption} 
\title{Few-Shot One-Class Classification via Meta-Learning}
\author{

  Ahmed Frikha \textsuperscript{\rm 1, 2, 4},
  Denis Krompaß \textsuperscript{\rm 1, 2},
  Hans-Georg K\"opken \textsuperscript{\rm 3},
  Volker Tresp \textsuperscript{\rm 2, 4}
  
}
\affiliations{

    \textsuperscript{\rm 1}Siemens AI Lab\\
    \textsuperscript{\rm 2}Siemens Technology\\
    \textsuperscript{\rm 3}Siemens Digital Industries\\
    \textsuperscript{\rm 4}Ludwig Maximilian University of Munich\\
    Munich, Germany\\

}

\begin{document}

\maketitle

\begin{abstract}
Although few-shot learning and one-class classification (OCC), i.e., learning a binary classifier with data from only one class, have been separately well studied, their intersection remains rather unexplored. Our work addresses the few-shot OCC problem and presents a method to modify the episodic data sampling strategy of the model-agnostic meta-learning (MAML) algorithm to learn a model initialization particularly suited for learning few-shot OCC tasks. This is done by explicitly optimizing for an initialization which only requires few gradient steps with one-class minibatches to yield a performance increase on class-balanced test data. We provide a theoretical analysis that explains why our approach works in the few-shot OCC scenario, while other meta-learning algorithms fail, including the unmodified MAML. Our experiments on eight datasets from the image and time-series domains show that our method leads to better results than classical OCC and few-shot classification approaches, and demonstrate the ability to learn unseen tasks from only few normal class samples. Moreover, we successfully train anomaly detectors for a real-world application on sensor readings recorded during industrial manufacturing of workpieces with a CNC milling machine, by using few normal examples. Finally, we empirically demonstrate that the proposed data sampling technique increases the performance of more recent meta-learning algorithms in few-shot OCC and yields state-of-the-art results in this problem setting.
\end{abstract}

\section{Introduction}\label{introduction}

The anomaly detection (AD) task \cite{chandola2009anomaly, aggarwal2015outlier} consists in differentiating between normal and abnormal data samples. AD applications are common in various domains that involve different data types, including medical diagnosis \cite{prastawa2004brain}, cybersecurity \cite{garcia2009anomaly} and quality control in industrial manufacturing \cite{scime2018anomaly}. Due to the rarity of anomalies, the data underlying AD problems exhibits high class-imbalance. Therefore, AD problems are usually formulated as one-class classification (OCC) problems \cite{moya1993one}, where either only a few or no anomalous data samples are available for training the model \cite{khan2014one}. While most of the developed approaches \cite{khan2014one} require a substantial amount of normal data to yield good generalization, in many real-world applications, e.g., in industrial manufacturing, only small datasets are available. Data scarcity can have many reasons: data collection itself might be expensive, e.g., in healthcare, or happens only gradually, such as in a cold-start situation, or the domain expertise required for annotation is scarce and expensive. 

To enable learning from few examples, viable approaches \cite{lake2011one, ravi2016optimization, finn2017model} relying on meta-learning \cite{schmidhuber1987evolutionary} have been developed. However, they rely on having examples from each of the task's classes, which prevents their application to OCC tasks.  While recent meta-learning approaches focused on the few-shot learning problem, i.e., learning to learn with few examples, we extend their use to the OCC problem, i.e., learning to learn with examples from only one class. To the best of our knowledge, the few-shot OCC (FS-OCC) problem has only been addressed in \cite{kozerawski2018clear, kruspe2019one} in the image domain.

Our contribution is fourfold: Firstly, we show that classical OCC approaches fail in the few-shot data regime. Secondly, we provide a theoretical analysis showing that classical gradient-based meta-learning algorithms do not yield parameter initializations suitable for OCC and that second-order derivatives are needed to optimize for such initializations. Thirdly, we propose a simple episode generation strategy to adapt any meta-learning algorithm that uses a bi-level optimization scheme to FS-OCC. Hereby, we first focus on modifying the model-agnostic meta-learning (MAML) algorithm \cite{finn2017model} to learn initializations useful for the FS-OCC scenario. The resulting One-Class MAML (OC-MAML) maximizes the inner product of loss gradients computed on one-class and class-balanced minibatches, hence maximizing the cosine similarity between these gradients. Finally, we demonstrate that the proposed data sampling technique generalizes beyond MAML to other metalearning algorithms, e.g., MetaOptNet \cite{lee2019meta} and Meta-SGD \cite{li2017meta}, by successfully adapting them to the understudied FS-OCC.

We empirically validate our approach on eight datasets from the image and time-series domains, and demonstrate its robustness and maturity for real-world applications by successfully testing it on a real-world dataset of sensor readings recorded during manufacturing of metal workpieces with a CNC milling machine. Furthermore, we outperform the concurrent work One-Way ProtoNets \citep{kruspe2019one} and achieve state-of-the-art performance in FS-OCC.

\section{Approach}

The primary contribution of our work is to propose a way to adapt meta-learning algorithms designed for class-balanced FS learning to the underexplored FS-OCC problem. In this section, as a first demonstration that meta-learning is a viable approach to this challenging learning scenario, we focus on investigating it on the MAML algorithm. MAML was shown to be a universal learning algorithm approximator \cite{finn2017metalearning}, i.e., it could approximate a learning algorithm tailored for FS-OCC. Later, we validate our methods on further meta-learning algorithms (Table \ref{results_further_meta_algos}).

\subsection{Problem statement} \label{problem_statement}

Our goal is to learn a one-class classification task using only a \emph{few} examples. In the following, we first discuss the unique challenges of the few-shot one-class classification (FS-OCC) problem. Subsequently, we discuss the formulation of the FS-OCC problem as a meta-learning problem.

To perform one-class classification, i.e., differentiate between in-class and out-of-class examples using only in-class data, approximating a \emph{generalized} decision boundary for the normal class is necessary. Learning such a class decision boundary in the few-shot regime can be especially challenging for the following reasons. On the one hand, if the model overfits to the few available datapoints, the class decision boundary would be too restrictive, which would prevent generalization to unseen examples. As a result, some normal samples would be predicted as anomalies. On the other hand, if the model overfits to the majority class, i.e., predicting almost everything as normal, the class decision boundary would overgeneralize, and out-of-class (anomalous) examples would not be detected. 

In the FS classification context, $N$-way $K$-shot learning tasks are used to test the learning procedure yielded by the meta-learning algorithm. An $N$-way $K$-shot classification task includes $K$ examples from \emph{each} of the $N$ classes that are used for learning this task, after which the trained classifier is tested on a disjoint set of data \cite{vinyals2016matching}. When the target task is an OCC task, only examples from one class are available for training, which can be viewed as a $1$-way $K$-shot classification task. To align with the anomaly detection problem, the available examples must belong to the normal (majority) class, which usually has a lower variance than the anomalous (minority) class. This problem formulation is a prototype for a practical use case where an application-specific anomaly detector is needed and only few normal examples are available.

\subsection{Model-Agnostic Meta-Learning} \label{MAML_section}

MAML is a meta-learning algorithm that we focus on adapting to the FS-OCC problem before validating our approach on further meta-learning algorithms (Table \ref{results_further_meta_algos}). MAML learns a model initialization that enables quick adaptation to unseen tasks using only few data samples. For that, it trains a model explicitly for few-shot learning on tasks $T_{i}$ coming from the same task distribution $p(T)$ as the unseen target task $T_{test}$. In order to assess the model's adaptation ability to \emph{unseen} tasks, the available tasks are divided into mutually disjoint task sets: one for meta-training $S^{tr}$, one for meta-validation $S^{val}$ and one for meta-testing $S^{test}$. Each task $T_{i}$ is divided into two disjoint sets of data, each of which is used for a particular MAML operation: $D^{tr}$ is used for adaptation and $D^{val}$ is used for validation, i.e., evaluating the adaptation. The adaptation of a model $f_{\theta}$ to a task $T_{i}$ consists in taking few gradient descent steps using \emph{few} datapoints sampled from $D^{tr}$ yielding $\theta^{'}_{i}$.

A good measure for the suitability of the initialization parameters $\theta$ for few-shot adaptation to a considered task $T_{i}$ is the loss $L^{val}_{T_{i}}(f_{\theta^{'}_{i}})$, which is computed on the validation set $D^{val}_{i}$ using the task-specific adapted model $f_{\theta^{'}_{i}}$. To optimize for few-shot learning, the model parameters $\theta$ are updated by minimizing the aforementioned loss across all meta-training tasks. This \emph{meta-update}, can be expressed as:
\begin{equation} 
\label{eq:outer-update}
  \theta \gets \theta - \beta \nabla_{\theta} \sum_{T_{i}\sim p(T)} L^{val}_{T_{i}}(f_{\theta^{'}_{i}}).
\end{equation}
Here $\beta$ is the learning rate used for the meta-update. To avoid overfitting to the meta-training tasks, model selection is done via validation using tasks from $S^{val}$. At meta-test time, the FS adaptation to unseen tasks from $S^{test}$ is evaluated. We note that, in the case of few-shot classification, $K$ datapoints from \emph{each} class are sampled from $D^{tr}$ for the adaptation, during training and testing.

\subsection{One-Class Model-Agnostic Meta-Learning} \label{approach-oc-maml}

\subsubsection{Algorithm.}

MAML learns a model initialization suitable for \emph{class-balanced} (CB) FS classification. To adapt it to FS-OCC, we aim to find a model initialization from which taking few gradients steps with a few one-class (OC) examples yields the same effect as doing so with a CB minibatch. We achieve this by adequately modifying the objective of the inner loop updates of MAML. Concretely, this is done by modifying the data sampling technique during meta-training, so that the class-imbalance rate (CIR) of the inner loop minibatches matches the one of the test task. 

\begin{algorithm}
\caption{Meta-training of OC-MAML}
\label{meta-training}
\begin{algorithmic}[1]
\REQUIRE $S^{tr}$: Set of meta-training tasks 
\REQUIRE $\alpha, \beta$: Learning rates
\REQUIRE $K, Q$: Batch size for the inner and outer updates
\REQUIRE $c$: CIR for the inner-updates
  \STATE Randomly initialize $\theta$
    \WHILE{not done}
      \STATE Sample batch of tasks $T_{i}$ from $S^{tr}$; $T_{i}=\{D^{tr},D^{val}\}$ 
      \FOR{\textbf{each} sampled $\emph{T}_{i}$}
        \STATE Sample $K$ examples $B$ from $D^{tr}$ such that CIR$=c$
        \STATE Initialize $\theta^{'}_{i} = \theta$
        \FOR{number of adaptation steps}
          \STATE Compute adapted parameters with gradient descent using $B$:
          $\theta^{'}_{i} = \theta^{'}_{i} - \alpha \nabla_{\theta^{'}_{i}} L^{tr}_{T_{i}}(f_{\theta^{'}_{i}})$
        \ENDFOR
        \STATE Sample $Q$ examples $B^{'}$ from $D^{val}$ w/ CIR$=50\%$
        \STATE Compute outer loop loss $L^{val}_{T_{i}}(f_{\theta^{'}_{i}})$ using $B^{'}$
      \ENDFOR
      \STATE Update $\theta$: $\theta \gets \theta - \beta \nabla_{\theta} \sum_{T_{i}} L^{val}_{T_{i}}(f_{\theta^{'}_{i}})$
    \ENDWHILE \\
\STATE \textbf{return} meta-learned parameters $\theta$
\end{algorithmic}
\end{algorithm}
MAML optimizes explicitly for FS adaptation by creating and using auxiliary tasks that have the same characteristic as the target tasks, in this case tasks that include only few datapoints for training. It does so by reducing the size of the batch used for the adaptation (via the hyperparameter $K$ \cite{finn2017metalearning}). Analogously, OC-MAML trains explicitly for quick adaptation to OCC tasks by creating OCC auxiliary tasks for meta-training. OCC problems are binary classification scenarios where only few or no minority class samples are available. In order to address both of theses cases, we introduce a hyperparameter ($c$) which sets the CIR of the batch sampled for the inner updates. Hereby, $c$ gives the percentage of the samples belonging to the minority (anomalous) class w.r.t. the total number of samples, e.g., setting $c=0\%$ means only majority class samples are contained in the data batch. We focus on this extreme case, where no anomalous samples are available for learning. In order to evaluate the performance of the adapted model on both classes, we use a \emph{class-balanced} validation batch $B^{'}$ for the meta-update. This way, we maximize the performance of the model in recognizing both classes after having \emph{seen} examples from only one class during adaptation. The OC-MAML meta-training is described in Algorithm 1, and the cross-entropy loss was used for $L$. At test time, the adaptation to an unseen task is done by applying steps 5-9 in Algorithm 1, starting from the meta-learned initialization.

We note that the proposed episode sampling strategy, i.e., training on a one-class batch then using the loss computed on a class-balanced validation batch to update the meta-learning strategy (e.g., model initialization), is applicable to any meta-learning algorithm that incorporates a bi-level optimization scheme (examples in Table \ref{results_further_meta_algos}).

\begin{figure}[h]
  \centering
  \includegraphics[width=0.4\textwidth]{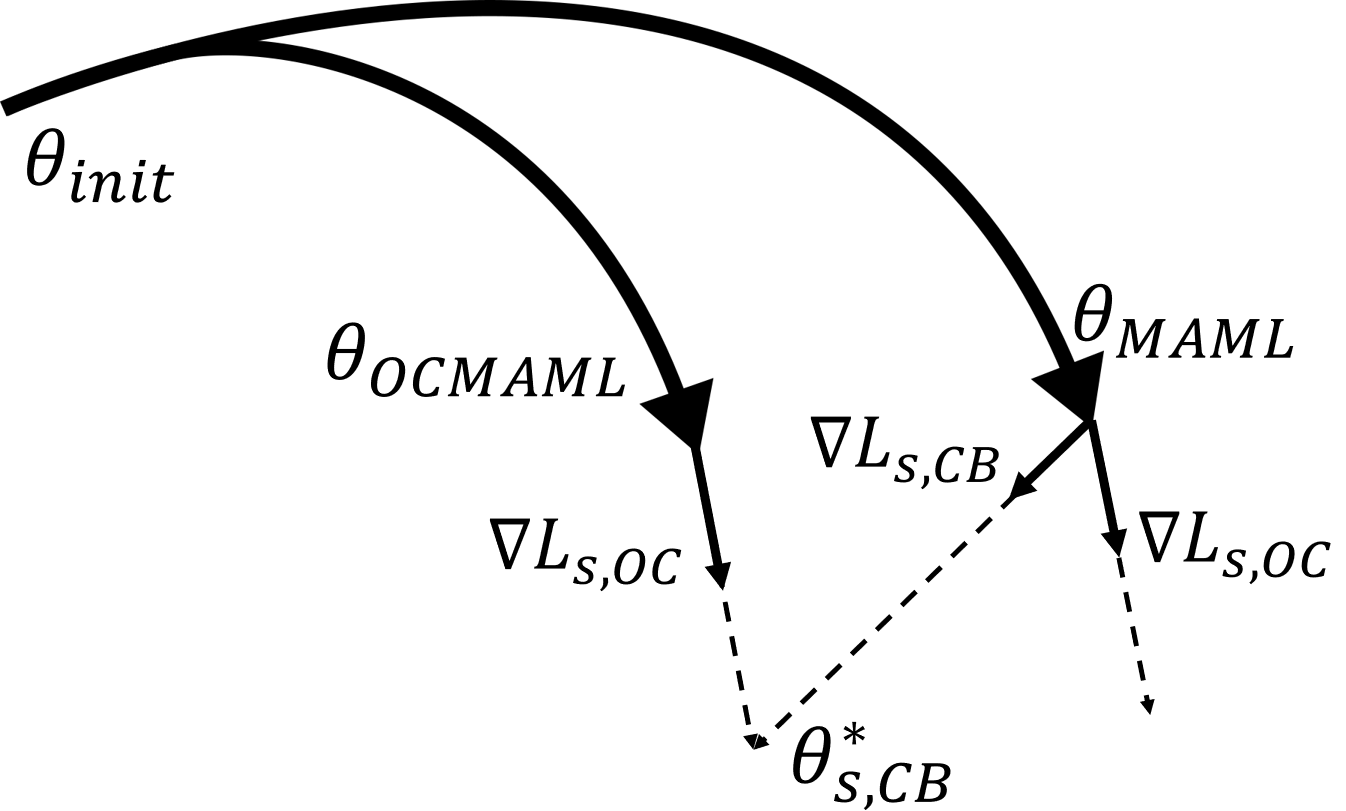}
  \caption{Adaptation to task $T_{s}$ from the model initializations yielded by OC-MAML and MAML}
  \label{fig:ocmaml_vs_maml}
\end{figure}

Using OCC tasks for adaptation during meta-training favors model initializations that enable a quick adaptation to OCC tasks over those that require CB tasks. The schematic visualization in Figure \ref{fig:ocmaml_vs_maml} shows the difference between the model initializations meta-learned by MAML and OC-MAML. Hereby, we consider the adaptation to an unseen binary classification task $T_{s}$. $\theta^{*}_{s,CB}$ denotes a local optimum of $T_{s}$. The parameter initializations yielded by OC-MAML and MAML are denoted by $\theta_\textit{OCMAML}$ and $\theta_\textit{MAML}$ respectively. When starting from the OC-MAML parameter initialization, taking a gradient step using an OC support set $D_{s,OC}$ (gradient direction denoted by $\nabla L_{s,OC}$), yields a performance increase on $T_{s}$ (by moving closer to the local optimum). In contrast, when starting from the parameter initialization reached by MAML, a class-balanced support set $D_{s,CB}$ (gradient direction denoted by $\nabla L_{s,CB}$) is required for a performance increase on $T_{s}$. 

\subsubsection{Theoretical Analysis: Why does OC-MAML work ?} \label{theoretical-analysis}

In this section we give a theoretical explanation of why OC-MAML works and why it is a more suitable approach than MAML for the FS-OCC setting. To address the latter problem, we aim to find a model parameter initialization, from which adaptation using few data examples from only \emph{one} class yields a good performance on both classes, i.e., good generalization to the class-balanced task. We additionally demonstrate that adapting first-order meta-learning algorithms, e.g., First-Order MAML (FOMAML) \cite{finn2017model} and Reptile \cite{nichol2018reptile}, to the OCC scenario as done in OC-MAML, does not yield initializations with the desired characteristics.

By using a Taylor series expansion the gradient used in the MAML update can be approximated to Equation \ref{gmaml} \cite{nichol2018reptile}, where the case with only 2 gradient-based updates is considered, i.e., one adaptation update on a minibatch (1), the support set including $K$ examples from $D^{tr}$, and one meta-update on a minibatch (2), the query set including $Q$ examples from $D^{val}$. We use the notation from \cite{nichol2018reptile}, where $\gbar_{i}$ and $\Hbar_{i}$ denote the gradient and Hessian computed on the $i^{th}$ minibatch at the initial parameter point $\phi_{1}$, and $\alpha$ the learning rate. Here it is assumed that the same learning rate is used for the adaptation and meta-updates.
\begin{equation} \label{gmaml}
\begin{split}
    g_\textit{MAML} &= \gbar_{2} - \alpha \Hbar_{2}\gbar_{1} - \alpha \Hbar_{1}\gbar_{2} + O(\alpha^2) \\
    &= \gbar_{2} - \alpha \frac{\partial (\gbar_{1}.\gbar_{2})}{\partial \phi_{1}} + O(\alpha^2)
\end{split}
\end{equation}
Equation \ref{gmaml} shows that MAML maximizes the inner product of the gradients computed on different minibatches \cite{nichol2018reptile}. Under the assumption of local linearity of the loss function (which is the case around small optimization steps), and when gradients from different minibatches have a positive inner product, taking a gradient step using one minibatch yields a performance increase on the other \cite{nichol2018reptile}. Maximizing the inner product leads to a decrease in the angle between the gradient vectors and thus to an increase in their cosine similarity. Hence, MAML optimizes for an initialization where gradients computed on \emph{small} minibatches have similar directions, which enables few-shot learning.  

Equation \ref{gmaml} is independent of the data strategy adopted and hence holds also for OC-MAML. However, in OC-MAML the minibatches $1$ and $2$ have different class-imbalance rates (CIRs), since the first minibatch includes examples from only one class and the second minibatch is class-balanced. So, it optimizes for increasing the inner product between a gradient computed on a one-class minibatch and a gradient computed on class-balanced data. Thus, OC-MAML optimizes for an initialization where gradients computed on one-class data have similar directions, i.e., a high inner product and therefore a high cosine similarity, to gradients computed on class-balanced data (Figure \ref{fig:ocmaml_vs_maml}). Consequently, taking one (or few) gradient step(s) with one-class minibatch(es) from such a parameter initialization results in a performance increase on class-balanced data. This enables one-class classification. In contrast, MAML uses only class-balanced data during meta-training, which leads to a parameter initialization that requires class-balanced minibatches to yield the same effect. When adapting to OCC tasks, however, only examples from one class are available. We conclude, therefore, that the proposed data sampling technique modifies MAML to learn parameter initializations that are more suitable for adapting to OCC tasks.  

A natural question is whether applying the same data sampling method to other gradient-based meta-learning algorithms would yield the same desired effect. We investigate this for First-Order MAML (FOMAML), a first-order approximation of MAML that ignores the second derivative terms and Reptile \cite{nichol2018reptile}, which is also a first-order meta-learning algorithm that learns an initialization that enables fast adaptation to test tasks using few examples from \emph{each} class. We refer to the versions of these algorithms adapted to the FS-OCC setting as OC-FOMAML and OC-Reptile. We note that for OC-Reptile, the first $N-1$ batches contain examples from only one class and the last ($N^{th}$) batch is class-balanced. The approximated FOMAML and Reptile gradients are given by Equations \ref{gfomaml} and \ref{greptile} \cite{nichol2018reptile}, respectively.

\begin{equation} \label{gfomaml}
    g_\textit{FOMAML}= \gbar_{2} - \alpha \Hbar_{2}\gbar_{1} + O(\alpha^2) 
\end{equation}
\begin{equation} \label{greptile}
    g_\textit{Reptile}= \gbar_{1} + \gbar_{2} - \alpha \Hbar_{2}\gbar_{1} + O(\alpha^2)
\end{equation} \\
We note that these equations hold also for OC-FOMAML and OC-Reptile. By taking the expectation over minibatch sampling $\mathbb{E}_{\tau,1,2}$ for a task $\tau$ and two \emph{class-balanced} minibatches $1$ and $2$, it is established that $\mathbb{E}_{\tau,1,2}[\Hbar_{1}\gbar_{2}] = \mathbb{E}_{\tau,1,2}[\Hbar_{2}\gbar_{1}]$ \cite{nichol2018reptile}. Averaging the two sides of the latter equation results in
\begin{equation} \label{innerproduct}
\begin{split}
    \mathbb{E}_{\tau,1,2}[\Hbar_{2}\gbar_{1}] &= \frac{1}{2}\mathbb{E}_{\tau,1,2}[\Hbar_{1}\gbar_{2} + \Hbar_{2}\gbar_{1}] \\
    &= \frac{1}{2}\mathbb{E}_{\tau,1,2}[\frac{\partial (\gbar_{1}.\gbar_{2})}{\partial \phi_{1}}].
\end{split}
\end{equation}

Equation \ref{innerproduct} shows that, FOMAML and Reptile, like MAML, in expectation optimize for increasing the inner product of the gradients computed on different minibatches with the \emph{same} CIR. However, when the minibatches $1$ and $2$ have different CIRs, which is the case for OC-FOMAML and OC-Reptile, $\mathbb{E}_{\tau,1,2}[\Hbar_{1}\gbar_{2}] \neq \mathbb{E}_{\tau,1,2}[\Hbar_{2}\gbar_{1}]$ and therefore $\mathbb{E}_{\tau,1,2}[\Hbar_{2}\gbar_{1}] \neq \frac{1}{2}\mathbb{E}_{\tau,1,2}[\frac{\partial (\gbar_{1}.\gbar_{2})}{\partial \phi_{1}}]$. Hence, despite using the same data sampling method as OC-MAML, OC-FOMAML and OC-Reptile do \emph{not} explicitly optimize for increasing the inner product, and therefore the cosine similarity, between gradients computed on one-class and class-balanced minibatches. The second derivative term $\Hbar_{1}\gbar_{2}$ is, thus, necessary to optimize for an initialization from which performance increase on a class-balanced task is yielded by taking few gradient steps using one class data.

\section{Related works} \label{related-works}

Our proposed method addresses the FS-OCC problem, i.e., solving binary classification problems using only \emph{few} datapoints from only \emph{one} class. To the best of our knowledge, this problem was only addressed in \cite{kozerawski2018clear} and \cite{kruspe2019one}, and exclusively in the image data domain. In \cite{kozerawski2018clear} a feed-forward neural network is trained on ILSVRC 2012 to learn a transformation from feature vectors, extracted by a CNN pre-trained on ILSVRC 2014 \cite{ILSVRC15}, to SVM decision boundaries. At test time, an SVM boundary is inferred by using one image of one class from the test task which is then used to classify the test examples. This approach is specific to the image domain since it relies on the availability of very large, well annotated datasets and uses data augmentation techniques specific to the image domain, e.g., mirroring. Meta-learning algorithms offer a more general approach to FS-OCC since they are data-domain-agnostic, and do not require a pre-trained feature extraction model, which may not be available for some data domains, e.g., sensor readings. 

The concurrent work One-Way ProtoNets \cite{kruspe2019one} adapts ProtoNets \cite{snell2017prototypical} to address FS-OCC by using $0$ as a prototype for the \emph{null} class, i.e., non-normal examples, since the embedding space is $0$-centered due to using batch normalization (BN) \cite{ioffe2015batch} as the last layer. Given the embedding of a query example, its distance to the normal-class prototype is compared to its norm. This method constraints the model architecture by requiring the usage of BN layers. We propose a model-architecture agnostic data sampling technique to adapt meta-learning algorithms to the FS-OCC problem. The resulting meta-learning algorithms substantially outperform One-Way ProtoNets \cite{kruspe2019one} (Table \ref{results_further_meta_algos}).

\subsection{Class-balanced few-shot classification}

Meta-learning approaches for FS classification approaches may be broadly categorized in 2 categories. Optimization-based approaches aim to learn an optimization algorithm \cite{ravi2016optimization} and/or a parameter initialization \cite{finn2017model, nichol2018reptile}, learning rates \cite{li2017meta}, an embedding network \cite{lee2019meta} that are tailored for FS learning. Metric-based techniques learn a metric space where samples belonging to the same class are close together, which facilitates few-shot classification \cite{Koch2015SiameseNN, vinyals2016matching, snell2017prototypical, Sung_2018_CVPR, oreshkin2018tadam, lee2019meta}. Hybrid methods \cite{rusu2018metalearning, lee2018gradient} combine the advantages of both categories. Prior meta-learning approaches to FS classification addressed the \emph{N}-way \emph{K}-shot classification problem described in Section \ref{problem_statement}, i.e they require examples from \emph{each} class of the test tasks. We propose a method to adapt meta-learning algorithm to the \emph{1}-way \emph{K}-shot scenario, where only few examples from \emph{one} class are available.

\subsection{One-class classification}

Classical OCC approaches rely on SVMs \cite{scholkopf2001estimating, tax2004support} to distinguish between normal and abnormal samples. Hybrid approaches combining SVM-based techniques with feature extractors were developed to compress the input data in lower dimensional representations \cite{Xu_2015, erfani2016high, andrews2016transfer}. Fully deep methods that jointly perform the feature extraction step and the OCC step have also been developed \cite{ruff2018deep}. Another category of approaches to OCC uses the reconstruction error of antoencoders \cite{hinton2006reducing} trained with only normal examples as an anomaly score \cite{hawkins2002outlier, an2015variational, chen2017outlier}. Yet, determining a decision threshold for such an anomaly score requires labeled data from both classes. Other techniques rely on GANs \cite{goodfellow2014generative} to perform OCC \cite{schlegl2017unsupervised, ravanbakhsh2017abnormal, sabokrou2018adversarially}. The aforementioned hybrid and fully deep approaches require a considerable amount of data from the OCC task to train the typically highly parametrized feature extractors specific to the normal class, and hence fail in the scarce data regime (Table \ref{results_all}).

\section{Experimental evaluation}

The conducted experiments \footnote{Code available under \url{https://github.com/AhmedFrikha/Few-Shot-One-Class-Classification-via-Meta-Learning}} use some modules of the pyMeta library \cite{spigler2019meta} and aim to address the following key questions: $(a)$ How do meta-learning-based approaches using the proposed episode sampling technique perform compared to classical OCC approaches in the few-shot (FS) data regime? $(b)$ Do our theoretical findings (Section \ref{theoretical-analysis}) about the differences between the MAML and OC-MAML initializations hold in practice? $(c)$ Does the proposed episode sampling strategy to adapt MAML to the FS-OCC setting yield the expected performance increase and does this hold for further meta-learning algorithms?

\subsection{Baselines and Datasets} \label{baselines-datasets}

We compare OC-MAML, with the classical OCC approaches One-Class SVM (OC-SVM) \cite{scholkopf2001estimating} and Isolation Forest (IF) \cite{liu2008isolation} (Question $(a)$), which we fit to raw features and embeddings of the support set of the test task. Here, we explore two types of embedding networks which are trained on the meta-training tasks as follows: one is trained in a Multi-Task-Learning (MTL) \cite{caruana1997multitask} setting using one-class-vs-all tasks and the other trained using the "Finetune" baseline (FB) \cite{DBLP:journals/corr/abs-1903-03096}. i.e., using multi-class classification on all classes available. 

Moreover, we compare first-order (FOMAML and Reptile) and second-order (MAML) class-balanced meta-learning algorithms to their adapted versions to the OCC scenario, i.e., OC-FOMAML and OC-Reptile and OC-MAML (Question $(b)$). Finally, we compare MetaOptNet \cite{lee2019meta} and meta-SGD \cite{li2017meta} to their one-class counterparts that use our sampling strategy (Question $(c)$). We conducted a hyperparameter search for each baseline separately and used the best performing setting for our experiments. We evaluate our approach on 8 datasets from the image and time-series data domains, including two synthetic time-series (STS) datasets that we propose as a benchmark for FS-OCC on time-series, and a real-world sensor readings dataset of CNC Milling Machine Data (CNC-MMD). To adapt the image datasets to the OCC scenario, we create binary classification tasks, where the normal class is one class of the initial dataset and the anomalous class contains examples from \emph{multiple} other classes.

\subsection{Results and Discussion} \label{results}
In this section, we first discuss the performance of classical OCC approaches and the meta-learning algorithms in the FS-OCC problem setting, as well as the impact of the proposed data sampling strategy. Subsequently, we demonstrate the maturity of our approach on a real-world dataset. Thereafter, we further confirm our theoretical analysis with empirical results of cosine similarity between gradients. Finally, we show the generalizability of our sampling technique to further meta-learning algorithms beyond MAML, and compare the resulting algorithms to One-Way ProtoNets.

\begin{table*}[h]
\vskip 0.1in
\begin{center}
\begin{tabular}{l | l l l l | l l l l}
\hline
\multicolumn{1}{c|}{Adaptation set size} &\multicolumn{4}{c|}{$K=2$} &\multicolumn{4}{c}{$K=10$}\\
\hline
Model $\backslash$ Dataset  &MIN &Omn &MNIST &Saw &MIN &Omn &MNIST &Saw  \\
\hline
FB           &$50.0$ &$50.6$ &$56.5$ &$50.0$        &$50.0$ &$51.2$ &$50.3$ &$50.0$  \\
MTL          &$50.0$ &$50.0$ &$49.7$ &$50.0$        &$50.2$ &$50.0$ &$45.3$ &$50.0$  \\
OC-SVM       &$50.2$ &$50.6$ &$51.2$ &$50.1$      &$51.2$ &$50.4$ &$53.6$ &$50.5$  \\
IF           &$50.0$ &$50.0$ &$50.0$ &$50.0$        &$50.7$ &$50.0$ &$50.9$ &$49.9$  \\
FB + OCSVM   &$50.0$ &$50.0$ &$55.5$ &$50.4$        &$51.4$ &$58.0$ &$86.6$ &$58.3$  \\
FB + IF      &$50.0$ &$50.0$ &$50.0$ &$50.0$        &$50.0$ &$50.0$ &$76.1$ &$51.5$  \\
MTL + OCSVM  &$50.0$ &$50.0$ &$50.0$ &$50.0$        &$50.0$ &$50.1$ &$53.8$ &$86.9$  \\
MTL + IF     &$50.0$ &$50.0$ &$50.0$ &$50.0$        &$50.0$ &$55.7$ &$84.2$ &$64.0$  \\

\hline


Reptile           &$51.6$ &$56.3$         &$71.1$ &$69.1$            &$57.1$ &$76.3$        &$89.8$ &$81.6$  \\
FOMAML          &$53.3$ &$78.8$            &$80.7$ &$75.1$           &$59.5$ &$93.7$         &$91.1$ &$80.2$  \\
MAML            &$62.3$ &$91.4$             &$85.5$ &$81.1$         &$65.5$ &$96.3$           &$92.2$ &$86$  \\
OC-Reptile           &$51.9$ &$52.1$         &$51.3$ &$51.6$              &$53.2$ &$51$             &$51.4$ &$53.2$  \\
OC-FOMAML       &$55.7$ &$74.7$          &$79.1$ &$58.6$         &$66.1$ &$87.5$         &$91.8$ &$73.2$  \\
OC-MAML (ours)      &$\bf69.1$ &$\bf96.6$         &$\bf88$ &$\bf96.6$      &$\bf76.2$ &$\bf97.6$          &$\bf95.1$ &$\bf95.7$\\
\end{tabular}
\end{center}
\caption{Accuracies (in $\%$) computed on the class-balanced test sets of the test tasks of MiniImageNet (MIN), Omniglot (Omn), MT-MNIST with $T_{test}=T_{0}$ and STS-Sawtooth (Saw).}
\label{results_all}
\end{table*}

Table \ref{results_all} shows the results averaged over 5 seeds of the classical OCC approaches (Top) and the meta-learning approaches, namely MAML, FOMAML, Reptile and their one-class versions (Bottom), on 3 image datasets and on the STS-Sawtooth dataset. For the meta-learning approaches, models were trained with and without BN layers and the results of the best architecture were reported for each dataset. The results of all the methods on the other 8 MT-MNIST task-combinations and on the STS-Sine dataset, are consistent with the results in Table \ref{results_all}.

While classical OCC methods yield chance performance in almost all settings, OC-MAML achieves very high results, consistently outperforming them across all datasets and on both support set sizes. Likewise, we observe that OC-MAML consistently outperforms the class-balanced and one-class versions of the meta-learning algorithms in all the settings, showing the benefits of our modification to MAML. 

Moreover, OC-FOMAML and OC-Reptile yield poor results, especially without BN, confirming our 
findings (Section \ref{theoretical-analysis}) that adapting first-order meta-learning algorithms to the OCC setting does not yield the desired effect. We found that using BN yields a substantial performance increase on the 3 image datasets and explain that by the gradient orthogonalizing effect of BN \cite{suteu2019regularizing}. In fact, gradient orthogonalization reduces interference between gradients computed on one-class and class-balanced batches. OC-MAML achieves high performance even without BN, as it reduces interference between these gradients by the means of its optimization objective (Section \ref{theoretical-analysis}).

Several previous meta-learning approaches, e.g., MAML \citep{finn2017model}, were evaluated in a transductive setting, i.e., the model classifies the whole test set at once which enables sharing information between test examples via BN \cite{nichol2018reptile}. In anomaly detection applications, the CIR of the encountered test set batches, and therefore the statistics used in BN layers, can massively change depending on the system behavior (normal or anomalous). Hence, we evaluate all methods in a non-transductive setting: we compute the statistics of all BN layers using the few one-class adaptation examples and use them for predictions on test examples. This is equivalent to classifying each test example separately. We also use this method during meta-training. We note that the choice of the BN scheme heavily impacts the performance of several meta-learning algorithms \citep{bronskill2020tasknorm}.

\subsubsection{Validation on the CNC-milling real-world dataset.}

We validate OC-MAML on the industrial sensor readings dataset CNC-MDD and report the results in Table \ref{tab:results_CNC-MMD}. We compute F1-scores for evaluation since the test sets are class-imbalanced. Depending on the type of the target milling operation (e.g., roughing), tasks created from \emph{different} operations from the same type are used for meta-training. OC-MAML consistently achieves high F1-scores between $80\%$ and $95.9\%$ across the 6 milling processes. The high performance on the minority class, i.e., in detecting anomalous data samples, is reached by using only $K=10$ non-anomalous examples ($c=0\%$). These results show that OC-MAML yielded a parameter initialization suitable for learning OCC tasks in the time-series data domain and the maturity of this method for industrial real-world applications. Due to the low number of anomalies, it is not possible to apply MAML with the standard sampling, which would require $K$ anomalous examples in the inner loop during meta-training. With OC-MAML, the few anomalies available are only used for the outer loop updates. We note that despite the high class-imbalance in the data of the meta-training processes, class-balanced query batches were sampled for the outer loop updates. This can be seen as an under-sampling of the majority class.
\begin{table}[h]
\begin{center}
\begin{tabular}{l | l | l | l | l | l}
\hline 
\multicolumn{1}{c|}{$F_{1}$}  &\multicolumn{1}{c|}{$F_{2}$} &\multicolumn{1}{c|}{$F_{3}$} &\multicolumn{1}{c|}{$F_{4}$} &\multicolumn{1}{c|}{$R_{1}$} &\multicolumn{1}{c}{$R_{2}$}\\ 
\hline 
$80.0 \%$    &$89.6 \%$  &$95.9 \%$  &$93.6 \%$  &$85.3 \%$ &$82.6 \%$\\
\end{tabular}
\end{center}
\caption{OC-MAML F1-scores, averaged over 150 tasks sampled from the test operations, on finishing ($F_{i}$) and roughing ($R_{j}$) operations of the real-world CNC-MMD dataset, with only $K=10$ normal examples ($c=0\%$).}
\label{tab:results_CNC-MMD}
\end{table}
\subsubsection{Cosine similarity analysis.}
\begin{table}[h]
\begin{center}
\begin{tabular}{ l  l  l  l  l }
\hline
Model $\backslash$ Dataset &MIN &Omn &MNIST &Saw   \\
\hline
Reptile           &$0.05$  &$0.02$    &$0.16$   &$0.02$   \\ 
FOMAML            &$0.13$    &$0.14$    &$0.31$   &$-0.02$\\
MAML              &$0.28$    &$0.16$    &$0.45$   &$0.01$\\
OC-Reptile        &$0.09$      &$0.05$      &$-0.09$     &$0.03$\\
OC-FOMAML         &$0.26$    &$0.12$    &$0.36$   &$0.07$\\
OC-MAML    &$\bf0.42$ &$\bf0.23$ &$\bf0.47$   &$\bf0.92$\\
\end{tabular}
\end{center}
\caption{Cosine similarity between the gradients of one-class and class-balanced minibatches averaged over test tasks of MiniImageNet, Omniglot, MT-MNIST and STS-Sawtooth.}
\label{results_cosine}
\end{table}

\begin{table*}[t]
\begin{center}
\begin{tabular}{l | c c c | c c c}
\hline
\multicolumn{1}{c|}{Support set size} &\multicolumn{3}{c|}{$K=2$} &\multicolumn{3}{c}{$K=10$}\\
\hline
Model $\backslash$ Dataset  &MIN &CIFAR-FS &FC100 &MIN &CIFAR-FS &FC100\\
\hline
MAML           &$62.3$      &$62.1$   &$55.1$          &$65.5$       &$69.1$ &$61.6$  \\ 
OC-MAML (ours) &$\bf69.1$    &$\bf70$ &$\bf59.9$       &$\bf76.2$  &$\bf79.1$ &$\bf65.5$  \\
\hline
MetaOptNet           &$50$           &$56$      &$51.2$          &$56.6$          &$74.8$ &$53.3$  \\
OC-MetaOptNet (ours)&$\bf51.8$     &$\bf56.3$ &$\bf52.2$        &$\bf67.4$       &$\bf75.5$ &$\bf59.9$  \\
\hline
MetaSGD           &$65$     &$58.4$ &$55$              &$73.6$                &$71.3$ &$61.3$  \\ 
OC-MetaSGD (ours) &$\bf69.6$     &$\bf71.4$ &$\bf60.3$     &$\bf75.8$   &$\bf77.8$ &$\bf64.3$  \\
\hline
One-Way ProtoNets \citep{kruspe2019one}  &$67$   &$70.9$ &$56.9$    &$74.4$   &$76.7$ &$62.1$  \\
\end{tabular}
\end{center}
\caption{Test accuracies (in $\%$) computed on the class-balanced test sets of the test tasks of MiniImageNet (MIN), CIFAR-FS and FC100 after using a one-class support set for task-specific adaptation}
\label{results_further_meta_algos}
\end{table*}
We would like to directly verify that OC-MAML maximizes the inner product, and therefore the cosine similarity, between the gradients of one-class and class-balanced batches of data, while the other meta-learning baselines do not (Figure \ref{fig:ocmaml_vs_maml}, Section \ref{theoretical-analysis}). For this, we use the initialization meta-learned by each algorithm to compute the loss gradient of $K$ normal examples and the loss gradient of a disjoint class-balanced batch. We use the best performing initialization for each meta-learning algorithm and compute the cosine similarities using on test tasks.

We report the mean cosine similarity on 3 image datasets and one time-series dataset in Table \ref{results_cosine}. The significant differences in the mean cosine similarity found between OC-MAML and the other meta-learning algorithms consolidate our theoretical findings from section \ref{theoretical-analysis}.

\subsubsection{Applicability to further metalearning algorithms and comparison to One-Way ProtoNets.}

To investigate whether the benefits of our sampling strategy generalize to further meta-learning algorithms beyond MAML, we apply it to MetaOptNet \cite{lee2019meta} and Meta-SGD \cite{li2017meta}. Like MAML, these algorithms use a bi-level optimization scheme (inner and outer loop optimization) to perform few-shot learning. This enables the application of our proposed data strategy which requires two sets of data with different CIRs to be used. We refer to the OC versions of these algorithms as OC-MetaOptNet and OC-MetaSGD.

MetaOptNet trains a representation network to extract feature embeddings that generalize well in the FS regime when fed to linear classifiers, e.g., SVMs. For that, a differentiable quadratic programming (QP) solver \cite{amos2017optnet} is used to fit the SVM \cite{lee2019meta} (inner loop optimization). The loss of the fitted SVM on a held-out validation set of the same task is used to update the representation network (outer loop optimization). Since solving a binary SVM requires examples from both classes and our sampling strategy provides one-class examples in the inner loop, we use an OC-SVM \cite{scholkopf2000support} classifier instead. The embeddings extracted for few normal examples by the representation network are used to fit the OC-SVM, which is then used to classify the class-balanced validation set and to update the embedding network, analogously to the class-balanced scenario. To fit the OC-SVM, we solve its dual problem \cite{scholkopf2000support} using the same differentiable quadratic programming (QP) solver \cite{amos2017optnet} used to solve the multi-class SVM in \cite{lee2019meta}. The ResNet-12 architecture is used for the embedding network. We use the meta-validation tasks to tune the OC-SVM hyperparameters. 

Meta-SGD meta-learns an inner loop learning rate for each model parameter besides the initalization. Our episode sampling method is applied as done for MAML. Unlike the class-balanced MetaSGD, the meta-learning optimization assigns negative values to some parameter-specific learning rates to counteract overfitting to the majority class, which leads to performing gradient ascent on the adaptation loss. To prevent this, we clip the learning rates between 0 and 1.

Table \ref{results_further_meta_algos} shows that applying the proposed sampling technique to MetaOptNet and Meta-SGD results in a significant accuracy increase in FS-OCC on the MiniImageNet, CIFAR-FS and FC100 datasets. Eventhough MetaOptNet substantially outperforms MAML and Meta-SGD in the class-balanced case \cite{lee2019meta}, it fails to compete in the FS-OCC setting, suggesting that meta-learning a suitable initialization for the classifier is important in this scenario. 

Finally, we compare to One-Way ProtoNets \footnote{We re-implemented One-Way ProtoNets to conduct the experiments, since the code from the original paper was not made public.} and find that OC-MAML and OC-MetaSGD significantly outperform it on all three datasets. The poorer performance of One-Way ProtoNets and OC-MetaOptNet could be explained by the absence of a mechanism to adapt the feature extractor (the convolutional layers) to the unseen test tasks. OC-MAML and OC-MetaSGD finetune the parameters of the feature extractor by the means of gradient updates on the few normal examples from the test task. We conducted experiments using 5 different seeds and present the average in Table \ref{results_further_meta_algos}.

\section{Conclusion}
This work addressed the novel and challenging problem of few-shot one-class classification (FS-OCC). We proposed an episode sampling technique to adapt meta-learning algorithms designed for class-balanced FS classification to FS-OCC. Our experiments on 8 datasets from the image and time-series domains, including a real-world dataset of industrial sensor readings, showed that our approach yields substantial performance increase on three meta-learning algorithms, significantly outperforming classical OCC methods and FS classification algorithms using standard sampling. Moreover, we provided a theoretical analysis showing that class-balanced gradient-based meta-learning algorithms (e.g., MAML) do not yield model initializations suitable for OCC tasks and that second-order derivatives are needed to optimize for such initializations. Future works could investigate an unsupervised approach to FS-OCC, as done in the class-balanced scenario \cite{hsu2018unsupervised}. \\

\bibliography{./references}

\newpage
\appendix

\section{Experimental Details} \label{exp-details}

In the following we provide details about the model architectures used. For MT-MNIST, we use the same 4-block convolutional architecture as used by \cite{hsu2018unsupervised} for their multi-class MNIST experiments. Each convolutional block includes a $3$ x $3$ convolutional layer with $32$ filters, a $2$ x $2$ pooling and a ReLU non-linearity. The same model architecture is used for the MiniImageNet experiments as done by \cite{ravi2016optimization}. For the Omniglot experiments, we use the same architecture used in \cite{finn2017model}. 

On the STS datasets, the model architecture used is composed of 3 modules, each including a $5$ x $5$ convolutional layer with $32$ filters, a $2$ x $2$ pooling and a ReLU non-linearity. The model architecture used for the CNC-MMD experiments is composed of 4 of these aforementioned modules, except that the convolutional layers in the last two modules include 64 filters. The last layer of all architectures is a linear layer followed by softmax. We note that in the experiments on the time-series datasets (STS and CNC-MMD) 1-D convolutional filters are used.

We conducted a hyperparameter grid search for each meta-learning algorithm separately. The hyperparameters with the most effect on the algorithm performance were identified and variated. These are: the inner and outer learning rates ($\alpha$ and $\beta$), the number of inner updates (adaptation steps). We also conducted a separate hyperparameter search for the case, where BN layers are used. Our results are averaged over 5 runs with different seeds, using the best hyperparameter values. For the outer learning rate $\beta$ we searched over the grid $\{0.1, 0.01, 0.001\}$ for all datasets. Regarding the inner learning rate ($\alpha$) we searched over the grids $\{0.1, 0.01, 0.001\}$ for MiniImageNet and the STS datasets, and $\{0.1, 0.05, 0.01\}$ for MT-MNIST and Omniglot. As for the number of adaptation steps, we search over the grids $\{1, 3, 5\}$ for MiniImageNet, MT-MNIST and Omniglot, and $\{1, 3, 5, 10\}$ for the STS datasets. 

For the meta-learning algorithms, including OC-MAML, we used vanilla SGD in the inner loop and the Adam optimizer \citep{kingma2014adam} in the outer loop, as done by \cite{finn2017model}. For (FO)MAML and OC-(FO)MAML, the size of the query set, also called outer loop minibatch, ($Q$), was set to $60$, $20$, $100$ and $50$, for MiniImageNet, Omnigot, MT-MNIST and the STS datasets respectively. Since the outer loop data is class-balanced, it includes $Q/2$ examples per class. Reptile uses the same batch size for all updates \citep{nichol2018reptile}. Hence, we set the outer loop minibatch size to be equal to the inner loop minibatch size, i.e., $Q=K$. The number of meta-training tasks used in each meta-training iteration (also called meta-batch size) was set to 8 for all datasets. 

The MTL and FB baselines were also trained with the Adam optimizer. Here, the batch size used is $32$ for all datasets, and the learning rate was set to $0.05$ for MiniImageNet and Omniglot and $0.01$ for MT-MNIST and STS. 

In the following we give the hyperparameters used for the real-world CNC-MMD dataset of industrial sensor readings. The outer learning rate ($\beta$) was set to $0.001$, the inner learning rate ($\alpha$) was set to $0.0001$, the number of adaptation steps was set to $5$ and the meta-batch size was set to $16$. Since the sizes and CIRS of the validation sets $D^{val}$ differ across the meta-training tasks in this dataset, we could not fix the outer loop size $Q$. Here we sample the $Q$ datapoints with the biggest possible size, under the constraint that these datapoints are class-balanced. The resulting $Q$ values are between $4$ and $16$, depending on the meta-training task.  

In the following, we provide details about the meta-training procedure adopted in the meta-learning experiments. We use disjoint sets of data for adaptation ($D^{tr}$) and validation ($D^{val}$) on the meta-training tasks, as it was empirically found to yield better final performance \citep{nichol2018reptile}. Hereby, the same sets of data are used in the OC-MAML and baseline experiments. In the MT-MNIST, Omniglot, MiniImageNet and STS experiments, the aforementioned sets of data are class-balanced. The sampling of the batch used for adaptation $B$ ensures that this latter has the appropriate CIR ($c=50\%$ for MAML, FOMAML and Reptile, and $c=c_{target}$ for OC-MAML, OC-FOMAML and OC-Reptile). For the one-class meta-learning algorithms, $c_{target}=0\%$, i.e., no anomalous samples of the target task are available, sothat only normal examples are sampled from  $D^{tr}$ during meta-training. In order to ensure that class-balanced and one-class meta-learning algorithms are exposed to the same data during meta-training, we move the anomalous examples from the adaptation set of data ($D^{tr}$) to the validation set of data ($D^{val}$). We note that this is only done in the experiments using one-class meta-learning algorithms.

During meta-training, meta-validation episodes are conducted to perform model selection. In order to mimic the adaptation to unseen FS-OCC tasks with CIR $c=c_{target}$ at test time, the CIR of the batches used for adaptation during meta-validation episodes is also set to $c=c_{target}$. We note that the hyperparameter $K$ denotes the total number of datapoints, i.e., batch size, used to perform the adaptation updates, and not the number of datapoints \emph{per class} as done by \cite{finn2017model}. Hence, a task with size $K=10$ and CIR $c=50\%$ is equivalent to a 2-way 5-shot classification task. \\

In the following, we provide details about the adaptation to the target task(s) and the subsequent evaluation. In the MT-MNIST and MiniImageNet experiments, we randomly sample 20 adaptation sets from the target task(s)' data, each including $K$ examples with the CIR corresponding to the experiment considered. After each adaptation episode conducted using one of these sets, the adapted model is evaluated on a disjoint class-balanced test set that includes 4,000 images for MT-MNIST and 600 for MiniImageNet. We note that the samples included in the test sets of the test tasks are not used nor for meta-training neither for meta-validation. This results in 20 and 400 (20 adaptation sets created from each of the 20 test classes) different test tasks for MT-MNIST and MiniImageNet, respectively. All the results presented give the mean over all adaptation episodes. Likewise, in the STS experiments, we evaluate the model on 10 different adaptation sets from each of the 5 test tasks. In the CNC-MMD experiments, the 30 tasks created from the target operation are used for adaptation and subsequent evaluation. For each of these target tasks, we randomly sample $K$ datapoints belonging to the normal class that we use for adaptation, and use the rest of the datapoints for testing. We do this 5 times for each target task, which results in 150 testing tasks. 
For MTL and FB baselines, as well as all the baseline combining these model with shallow models, i.e., IF and OC-SVM, we use the meta-validation task(s) for model choice, like in the meta-learning experiments. For the MTL baseline, for each validation task, we finetune a fully connected layer on top of the shared multi-task learned layers, as it is done at test time.

\section{Datasets and task creation procedures} \label{appendix-datasets}

In this Section we first provide general information about the datasets used in our experiments. Subsequently, we present more detailed information about the original datasets, the procedures adopted for creating OCC tasks, and the steps adopted to create the proposed STS datasets. 

We evaluate our approach on 8 datasets from the image and time-series data domains. From the image domain we use 4 few-shot learning benchmarks, namely MiniImageNet \cite{ravi2016optimization}, Omniglot \cite{Lake1332}, CIFAR-FS \cite{bertinetto2018meta} and FC100 \cite{oreshkin2018tadam} and 1 OCC benchmark dataset, the Multi-Task MNIST (MT-MNIST) dataset. To adapt the datasets to the OCC scenario, we create binary classification tasks, where the normal class contains examples from one class of the initial dataset and the anomalous class contains examples from \emph{multiple} other classes. We note that the class-balanced versions of the meta-learning baselines, e.g., MAML and Reptile, are trained with class-balanced data batches from such AD tasks in the inner loop of meta-training. We create 9 sub-datasets based on MNIST, where the meta-testing task of each consists in differentiating between a certain digit and the others, and the same ($10^{th}$) task for meta-validation in all sub-datasets. 

Since most of the time-series datasets for anomaly detection include data from only one domain and only one normal class, it is not possible them to the meta-learning problem formulation where several different tasks are required. Therefore, we create two synthetic time-series (STS) datasets, each including 30 synthetically generated time-series that underlie 30 different anomaly detection tasks. The time-series underlying the datasets are sawtooth waveforms (STS-Sawtooth) and sine functions (STS-Sine). We propose the STS-datasets as benchmark datasets for the few-shot (one-class) classification problem in the time-series domain and will publish them upon paper acceptance. 
Finally, we validate OC-MAML on a real-world anomaly detection dataset of sensor readings recorded during industrial manufacturing using a CNC milling machine. Various consecutive roughing and finishing operations (pockets, edges, holes, surface finish) were performed on ca. $100$ aluminium workpieces to record the CNC Milling Machine Data (CNC-MMD). The temporal dimension is handled using 1-D convolutions. 

In the following, we give details about all datasets, the task creation procedures adopted to adapt them to the OCC case, as well as the generation of the STS-datasets.

\textbf{Multi-task MNIST (MT-MNIST):} We derive 10 binary classification tasks from the MNIST dataset \citep{mnistCite}, where every task consists in recognizing one of the digits. This is a classical one-class classification benchmark dataset. For a particular task $T_{i}$, images of the digit $i$ are labeled as normal samples, while out-of-distribution samples, i.e., the other digits, are labeled as anomalous samples. We use 8 tasks for meta-training, 1 for meta-validation and 1 for meta-testing. Hereby, images of digits to be recognized in the validation and test tasks are not used as anomalies in the meta-training tasks. This ensures that the model is not exposed to normal samples from the test task during meta-training. Moreover, the sets of anomalous samples of the meta-training, meta-validation and meta-testing tasks are mutually disjoint. We conduct experiments on 9 MT-MNIST datasets, each of which involves a different target task ($T_{0}-T_{8}$). The task $T_{9}$ is used as a meta-validation task across all experiments. Each image has the shape $28$x$28$.

\textbf{Omniglot:} This dataset was proposed in \cite{Lake1332} and includes 20 instances of 1623 hand-written characters from 50 different alphabets. We generate our meta-training and meta-testing tasks based on the official data split \citep{Lake1332}, where 30 alphabets are reserved for training and 20 for evaluation. For each character class, we create a binary classification task, which consists in differentiating between this character and other characters from the same set (meta-training or meta-testing), i.e., the anomalous examples of a task $T_{i}$ are randomly sampled from the remaining characters. By removing 80 randomly sampled tasks from the meta-training tasks, we create the meta-validation tasks set. Each image has the shape $28$x$28$.

\textbf{MiniImageNet:} This dataset was proposed in \cite{ravi2016optimization} and includes 64 classes for training, 16 for validation and 20 for testing, and is a classical challenging benchmark dataset for few-shot learning. 600 images per class are available. To adapt it to the few-shot \emph{one-class} classification setting, we create 64 binary classification tasks for meta-training, each of which consists in differentiating one of the training classes from the others, i.e., the anomalous examples of a task $T_{i}$ are randomly sampled from the 63 classes with labels different from $i$. We do the same to create 16 meta-validation and 20 meta-testing tasks using the corresponding classes. Each image has the shape $84$x$84$x$3$. 

\textbf{CIFAR-FS:} This dataset was proposed in \cite{bertinetto2018meta} and includes 64 classes for training, 16 for validation and 20 for testing, derived from CIFAR-100, and is a benchmark dataset for few-shot learning. 600 images of size $32$x$32$x$3$ are available per class. To adapt it to the few-shot \emph{one-class} classification setting, we proceeded exactly as we did for miniImageNet (see above). 

\textbf{FC100:} This dataset was proposed in \cite{oreshkin2018tadam} and also includes 64 classes for training, 16 for validation and 20 for testing derived from CIFAR-100, and is a benchmark dataset for few-shot learning. However, in this dataset, the classes for training, validation and testing belong to different superclasses to minimize semantic overlap. This dataset contains 600 images of size $32$x$32$x$3$ per class. To adapt it to the few-shot \emph{one-class} classification setting, we proceeded exactly as we did for miniImageNet (see above). 

\textbf{Synthetic time-series (STS):} In order to investigate the applicability of OC-MAML to time-series (question $(c)$), we created two datasets, each including 30 synthetically generated time-series that underlie 30 different anomaly detection tasks. The time-series underlying the datasets are sawtooth waveforms (STS-Sawtooth) and sine functions (STS-Sine). Each time-series is generated with random frequencies, amplitudes, noise boundaries, as well as anomaly width and height boundaries. Additionally, the width of the rising ramp as a proportion of the total cycle is sampled randomly for the sawtooth dataset, which results in tasks having rising and falling ramps with different steepness values. The data samples of a particular task are generated by randomly cropping windows of length $128$ from the corresponding time-series. We generate 200 normal and 200 anomalous data examples for each task. For each dataset, we randomly choose 20 tasks for meta-training, 5 for meta-validation and 5 for meta-testing. We propose the STS-datasets as benchmark datasets for the few-shot one-class classification problem in the time-series domain, and will make them public upon paper acceptance.

In the following, we give details about the generation procedure adopted to create the STS-Sawtooth dataset. The same steps were conducted to generate the STS-Sine dataset. First, we generate the sawtooth waveforms underlying the different tasks by using the Signal package of the Scipy library \citep{scipy}. Thereafter, a randomly generated noise is applied to each signal. Subsequently, signal segments with window length $l=128$ are randomly sampled from each noisy signal. These represent the normal, i.e., non-anomalous, examples of the corresponding task. Then, some of the normal examples are randomly chosen, and anomalies are added to them to produce the anomalous examples. 

\begin{figure*}[h]
\caption{Exemplary normal (left) and anomalous (right) samples belonging to different tasks from the STS-Sawtooth (a and b) and the STS-Sine (c and d) datasets}
\label{fig:sts-signals}
\begin{center}
\includegraphics[width=14cm]{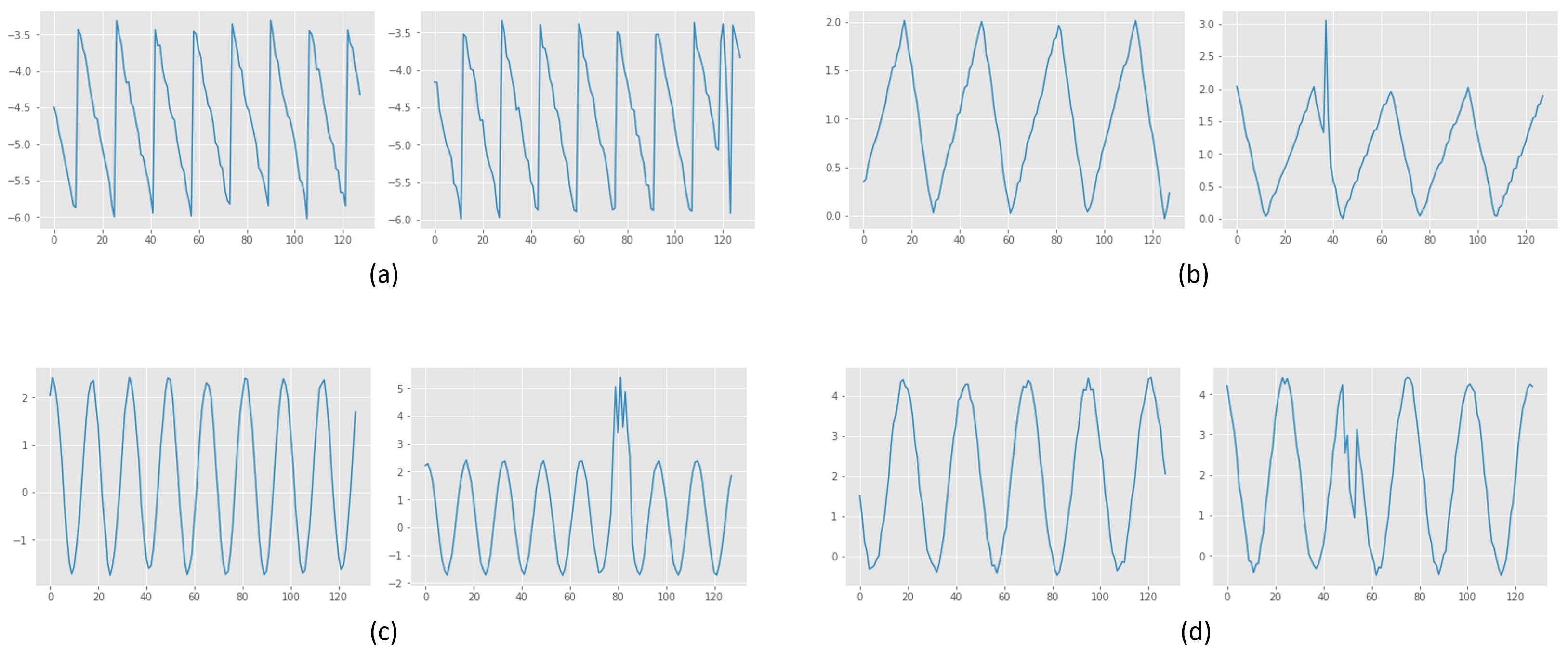}
\end{center}
\end{figure*}

Figure \ref{fig:sts-signals} shows exemplary normal and anomalous samples from the STS-Sawtooth and STS-Sine datasets. In order to increase the variance between the aforementioned synthetic signals underlying the different tasks, we randomly sample the frequency, i.e., the number of periods within the window length $l$, with which each waveform is generated, as well as the amplitude and the vertical position (see Figure \ref{fig:sts-signals}). For sawtooth waveforms, we also randomly sample the width of the rising ramp as a proportion of the total cycle between $0\%$ and $100\%$, for each task. Setting this value to $100\%$ and to $0\%$ produces sawtooth waveforms with rising and falling ramps, respectively. Setting it to $50\%$ corresponds to triangle waveforms. 

We note that the noise applied to the tasks are randomly sampled from \emph{task-specific} intervals, the boundaries of which are also randomly sampled. Likewise, the width and height of each anomaly is sampled from a random task specific-interval. Moreover, we generate the anomalies of each task, such that half of them have a height between the signal’s minimum and maximum (e.g., anomalies $(a)$ and $(d)$ in Figure \ref{fig:sts-signals}), while the other half can surpass these boundaries, i.e., the anomaly is higher than the normal signal’s maximum or lower than its minimum at least at one time step (e.g., anomalies $(b)$ and $(c)$ in Figure \ref{fig:sts-signals}). We note that an anomalous sample can have more than one anomaly.

We preprocess the data by removing the mean and scaling to unit variance. Hereby, only the available \emph{normal} examples are used for the computation of the mean and the variance. This means that in the experiments, where the target task's size $K=2$ and only normal samples are available $c=0\%$, only two examples are used for the mean and variance computation. We note that the time-series in Figure \ref{fig:sts-signals} are not preprocessed.

\textbf{CNC Milling Machine Data (CNC-MMD):} This dataset consists of ca. 100 aluminum workpieces on which various consecutive roughing and finishing operations (pockets, edges, holes, surface finish) are performed. The sensor readings which were recorded at a rate of 500Hz measure various quantities that are important for the process monitoring including the torques of the various axes.
Each run of machining a single workpiece can be seen as a multivariate time-series. We segmented the data of each run in the various operations performed on the workpieces. e.g., one segment would describe the milling of a pocket where another describes a surface finish operation on the workpiece. Since most manufacturing processes are highly efficient, anomalies are quite rare but can be very costly if undetected. For this reason, anomalies were provoked for 6 operations during manufacturing to provide a better basis for the analysis. Anomalies were provoked by creating realistic scenarios for deficient manufacturing. Examples are using a workpiece that exhibits deficiencies which leads to a drop in the torque signal or using rather slightly decalibrated process parameters which induced various irritations to the workpiece surface which harmed production quality. The data was labeled by domain experts from Siemens Digital Industries. It should be noted that this dataset more realistically reflects the data situation in many real application scenarios from industry where anomalies are rare and data is scarce and for this reason training models on huge class-balanced datasets is not an option.
  
For our experiments, we created 30 tasks per operation by randomly cropping windows of length $2048$ from the corresponding time-series of each operation. As a result, the data samples of a particular task $T_{i}$ cropped from a milling operation $O_{j}$ correspond to the same trajectory part of $O_{j}$, but to different workpieces. The task creation procedure ensures that at least two anomalous data samples are available for each task. The resulting tasks include between 15 and 55 normal samples, and between 2 and 4 (9 and 22) anomalous samples for finishing (roughing) operations. We validate our approach on all 6 milling operations in the case where only 10 samples belonging to the normal class ($K=10$, $c=0\%$) are available. Given the type of the target milling operation,e.g., finishing, we use the tasks from the other operations of the same type for meta-training. We note that the model is not exposed to any sample belonging to any task of the target operation during training. Each example has the shape $2048$x$3$.

We preprocess each of the three signals separately by removing the mean and scaling to unit variance, as done for the STS datasets. Likewise, only the available \emph{normal} examples are used for the computation of the mean and the variance.

\begin{figure*}[h]
\caption{Exemplary anomalous samples from a finishing (left) and a roughing (right) operations, where the anomalous time-steps are depicted in red.}
\label{fig:cnc-mmd-signals}
\begin{center}
\includegraphics[width=15cm]{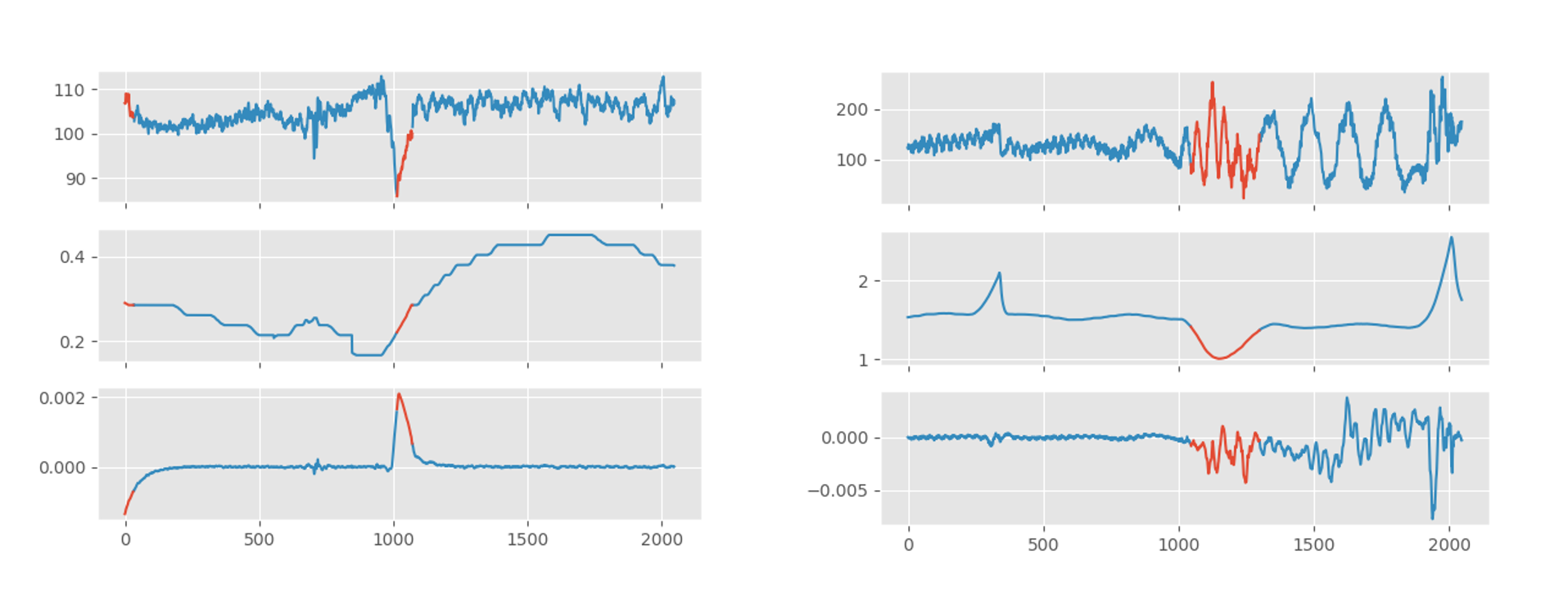}
\end{center}
\end{figure*}

Exemplary anomalous signals recorded from a finishing and a roughing operations are shown in Figure \ref{fig:cnc-mmd-signals}. These signals are not mean centered and scaled to unit variance. We note that we do not use the labels per time-step, but rather the label "anomalous" is assigned to each time-series that contains at least an anomalous time-step.

\section{Further Experimental Results} \label{appendix-exp-results}

In this section, we first present a more detailed overview of the experimental results on meta-learning algorithms, with and without using Batch Normalization (BN) layers in Table \ref{results_meta_complete}. Subsequently, we report the results of the experiments on the STS-Sine dataset (Tables \ref{results_occ_sts_sine} and \ref{results_meta_sts_sine}) and the 8 further MT-MNIST task combinations (Tables \ref{results_occ_mnist_1to4}, \ref{results_occ_mnist_5to8}, \ref{results_meta_mnist_1to4}, \ref{results_meta_mnist_5to8}). 

On the STS-Sine dataset and the 8 other MT-MNIST task combinations, we observe consistent results with the results from section \ref{results}. OC-MAML yields high performance across all datasets. We also note that none of the meta-learning baselines consistently yields high performance across all datasets, as it is the case for OC-MAML.

\begin{table*}[h]
\caption{Test accuracies (in $\%$) computed on the class-balanced test sets of the test tasks of MiniImageNet (MIN), Omniglot (Omn), MT-MNIST with $T_{test}=T_{0}$ and STS-Sawtooth (Saw). The results are shown for models without BN (top) and with BN (bottom), and give the average over 5 different seeds. One-class support sets ($c=0\%$) are used, unless otherwise specified.}
\label{results_meta_complete}
\vskip 0.1in
\begin{center}
\begin{tabular}{l | l l l l | l l l  l}
\hline
\multicolumn{1}{c|}{Support set size} &\multicolumn{4}{c|}{$K=2$} &\multicolumn{4}{c}{$K=10$}\\
\hline
Model $\backslash$ Dataset  &MIN &Omn &MNIST &Saw &MIN &Omn &MNIST &Saw\\
\hline
Reptile         &$50.2$ &$50$              &$71.1$ &$50.6$        &$50.2$ &$56.2$                 &$85.2$ &$72.8$  \\
FOMAML          &$50$ &$50.9$                &$80.7$ &$52.5$        &$50$ &$50.6$                 &$83.8$ &$50.4$  \\
MAML            &$51.4$ &$87.2$             &$80.7$ &$81.1$      &$50$ &$92.3$                    &$91.9$ &$72.9$  \\
OC-Reptile           &$50$ &$50.4$            &$50$ &$50$          &$50$ &$50.7$                      &$50$ &$50$  \\
OC-FOMAML       &$54.6$ &$52.2$             &$57.5$ &$55.9$        &$51$ &$53.2$              &$73.3$ &$58.9$  \\
OC-MAML (ours)      &$66.4$ &$95.6$ &$85.2$ &$\bf96.6$      &$73$ &$96.8$ &$\bf95.1$ &$\bf95.7$\\
\hline
Reptile (BN)           &$51.6$ &$56.3$         &$61.8$ &$69.1$            &$57.1$ &$76.3$        &$89.8$ &$81.6$  \\
FOMAML (BN)         &$53.3$ &$78.8$            &$80$ &$75.1$           &$59.5$ &$93.7$         &$91.1$ &$80.2$  \\
MAML (BN)            &$62.3$ &$91.4$             &$85.5$ &$51.7$         &$65.5$ &$96.3$           &$92.2$ &$86$  \\
OC-Reptile (BN)          &$51.9$ &$52.1$         &$51.3$ &$51.6$              &$53.2$ &$51$             &$51.4$ &$53.2$  \\
OC-FOMAML (BN)      &$55.7$ &$74.7$          &$79.1$ &$58.6$         &$66.1$ &$87.5$         &$91.8$ &$73.2$  \\
OC-MAML (BN) (ours)      &$\bf69.1$ &$\bf96.6$         &$\bf88$ &$51.3$      &$\bf76.2$ &$\bf97.6$          &$\bf95.1$ &$88.8$\\
\end{tabular}
\end{center}
\vskip -0.1in
\end{table*}

\begin{table}[h]
\caption{Test accuracies (in $\%$) computed on the class-balanced test sets of the test tasks of the STS-Sine dataset. One-class adaptation sets ($c=0\%$) are used, unless otherwise specified.}
\label{results_occ_sts_sine}
\begin{tabular}{| l | l | l |}
\hline
\multicolumn{1}{|c|}{Model $\backslash$ Adaptation set size} &\multicolumn{1}{c|}{$K=2$} &\multicolumn{1}{c|}{$K=10$}\\
\hline
FB ($c=50\%$)           &$68.9$ &$77.7$  \\
MTL ($c=50\%$)          &$64.5$ &$91.2$   \\
\hline
FB           &$73.8$ &$76.6$   \\
MTL          &$50$ &$50$  \\
OC-SVM       &$50.2$ &$51.3$   \\
IF           &$50$ &$49.9$   \\
FB + OCSVM   &$52.1$ &$65.3$   \\
FB + IF      &$50$ &$62.8$  \\
MTL + OCSVM  &$50$ &$51.9$   \\
MTL + IF     &$50$ &$64.7$   \\
OC-MAML (ours)      &$\bf99.9$ &$\bf99.9$   \\

\hline
\end{tabular}
\end{table}

\begin{table}[h]
\caption{Test accuracies (in $\%$) computed on the class-balanced test sets of the test tasks of the STS-Sine dataset. The results are shown for models without BN (top) and with BN (bottom). One-class adaptation sets ($c=0\%$) are used, unless otherwise specified.}
\label{results_meta_sts_sine}
\begin{tabular}{| l | l | l |}
\hline
\multicolumn{1}{|c|}{Model $\backslash$ Adaptation set size} &\multicolumn{1}{c|}{$K=2$} &\multicolumn{1}{c|}{$K=10$}\\
\hline
Reptile         &$52.5$ &$50.0$   \\
FOMAML          &$60.3$ &$52.1$         \\
MAML            &$99.6$ &$99.1$      \\
OC-Reptile           &$50.0$ &$50.0$       \\
OC-FOMAML       &$78.7$ &$58.1$          \\
OC-MAML (ours)  &$\bf99.9$ &$\bf99.9$   \\
\hline
Reptile (w. BN)           &$90.9$ &$98.6$         \\
FOMAML (w. BN)         &$90.8$ &$97.3$        \\
MAML (w. BN)            &$51.4$ &$99.0$     \\
OC-Reptile (w. BN)          &$52.6$ &$53.4$      \\
OC-FOMAML (w. BN)      &$78.8$ &$80.0$     \\
OC-MAML (w. BN) (ours)      &$50.5$ &$95.5$   \\
\hline

\end{tabular}
\end{table}

\begin{table*}[h]
\begin{center}
\caption{Test accuracies (in $\%$) computed on the class-balanced test sets of the test tasks of the MT-MNIST datasets with $T_{test}=T_{1-4}$. One-class adaptation sets ($c=0\%$) are used, unless otherwise specified.}
\label{results_occ_mnist_1to4}
\begin{tabular}{| l | l | l | l | l | l | l | l |  l | l | l | l | l |}
\hline
\multicolumn{1}{|c|}{Adaptation set size} &\multicolumn{4}{c|}{$K=2$} &\multicolumn{4}{c|}{$K=10$}\\
\hline
Model $\backslash$ Dataset  &$1$ &$2$ &$3$ &$4$ &$1$ &$2$ &$3$ &$4$  \\
\hline
FB ($c=50\%$)           &$78.8$ &$59.8$ &$66.7$ &$66.8$        &$91.9$   &$77.3$ &$79.9$ &$81.5$  \\
MTL ($c=50\%$)          &$64.9$ &$65$ &$59.5$ &$56.4$          &$91$ &$84.6$ &$84.4$ &$83.3$  \\
\hline
FB           &$53.7$ &$56$ &$50.7$ &$57.1$        &$53.6$ &$50.7$ &$50.2$ &$59$  \\
MTL          &$54$ &$46.8$ &$41.5$ &$52$        &$49.4$ &$49.6$ &$54.7$ &$46.1$  \\
OC-SVM       &$56.9$ &$51.5$ &$50.5$ &$51.8$      &$63.7$ &$50.2$ &$51.2$ &$51.5$  \\
IF           &$50$ &$50$ &$50$ &$50$        &$50.9$ &$50$ &$50.1$ &$50$  \\
FB + OCSVM   &$50.1$ &$53.2$ &$51.8$ &$56.1$        &$62.5$ &$70.5$ &$80.4$ &$89.8$  \\
FB + IF      &$50$ &$50$ &$50$ &$50$        &$54.3$ &$51.3$ &$77.7$ &$67.4$  \\
MTL + OCSVM  &$50$ &$50$ &$50$ &$50$        &$50.2$ &$52.8$ &$54.8$ &$50.7$  \\
MTL + IF     &$50$ &$50$ &$50$ &$50$        &$76.5$ &$75.5$ &$69.3$ &$74.4$  \\
OC-MAML (ours)      &$\bf87.1$ &$\bf86.3$         &$\bf86.8$ &$\bf85.9$      &$\bf92.5$ &$\bf92.4$     &$\bf91.7$ &$\bf92$\\
\hline
\end{tabular}
\end{center}
\end{table*}

\begin{table*}[h]
\begin{center}
\caption{Test accuracies (in $\%$) computed on the class-balanced test sets of the test tasks of the MT-MNIST datasets with $T_{test}=T_{5-8}$. One-class adaptation sets ($c=0\%$) are used, unless otherwise specified.}
\label{results_occ_mnist_5to8}
\begin{tabular}{| l | l | l | l | l | l | l | l |  l | l | l | l | l |}
\hline
\multicolumn{1}{|c|}{Adaptation set size} &\multicolumn{4}{c|}{$K=2$} &\multicolumn{4}{c|}{$K=10$}\\
\hline
Model $\backslash$ Dataset  &$1$ &$2$ &$3$ &$4$ &$1$ &$2$ &$3$ &$4$  \\
\hline
FB ($c=50\%$)           &$64.6$ &$69.8$ &$68.9$ &$62.9$        &$64.4$   &$83$ &$87.8$ &$72.8$  \\
MTL ($c=50\%$)          &$60.5$ &$71.4$ &$65$ &$60.6$          &$88.4$ &$91.4$ &$82$ &$79.1$  \\
\hline
FB           &$52.2$ &$66.5$ &$54.3$ &$53.8$        &$58.3$ &$63.5$ &$53.6$ &$50.1$  \\
MTL          &$48.5$ &$56.2$ &$51.1$ &$50.1$        &$49.9$ &$51.4$ &$48.5$ &$49.6$  \\
OC-SVM       &$51$ &$53.4$ &$53.9$ &$50.1$      &$50.5$ &$54$ &$54$ &$52.2$  \\
IF           &$50$ &$50$ &$50$ &$50$        &$50$ &$50.2$ &$49.8$ &$50.2$  \\
FB + OCSVM   &$52.2$ &$51.2$ &$50.5$ &$58$        &$86.2$ &$75$ &$84.5$ &$80$  \\
FB + IF      &$50$ &$50$ &$50$ &$50$        &$80.4$ &$87.2$ &$79.2$ &$71.4$  \\
MTL + OCSVM  &$50$ &$50$ &$50$ &$50$        &$51$ &$59.1$ &$71.3$ &$75.9$  \\
MTL + IF     &$50$ &$50$ &$50$ &$50$        &$50$ &$55.7$ &$84.2$ &$64$  \\
OC-MAML (ours)      &$\bf85.9$ &$\bf91.5$         &$\bf85.1$ &$\bf82.5$      &$\bf91.5$ &$\bf95.4$          &$\bf91.4$ &$\bf89.8$ \\

\hline
\end{tabular}
\end{center}
\end{table*}

\begin{table*}[h]
\begin{center}
\caption{Test accuracies (in $\%$) computed on the class-balanced test sets of the test tasks of the MT-MNIST datasets with $T_{test}=T_{1-4}$. The results are shown for models without BN (top) and with BN (bottom). One-class adaptation sets ($c=0\%$) are used, unless otherwise specified.}
\label{results_meta_mnist_1to4}
\begin{tabular}{| l | l | l | l | l | l | l | l |  l | l | l | l | l |}
\hline
\multicolumn{1}{|c|}{Adaptation set size} &\multicolumn{4}{c|}{$K=2$} &\multicolumn{4}{c|}{$K=10$}\\
\hline
Model $\backslash$ Dataset  &$1$ &$2$ &$3$ &$4$ &$1$ &$2$ &$3$ &$4$  \\
\hline
Reptile         &$67.1$ &$58.3$              &$57$ &$65.9$        &$82.4$ &$78.5$                 &$76.4$ &$81.8$  \\
FOMAML          &$76.3$ &$74.2$                &$74.9$ &$75.6$        &$82.3$ &$75.7$                 &$75.1$ &$80.8$  \\
MAML            &$78.1$ &$71.8$             &$77$ &$71.4$      &$88.8$ &$88.7$                    &$87.2$ &$86.6$  \\
OC-Reptile           &$50$ &$50$            &$50$ &$50$          &$50$ &$50$                      &$50$ &$50$  \\
OC-FOMAML       &$56.6$ &$52.6$             &$55.6$ &$50.1$        &$50.7$ &$50$              &$53.8$ &$64.3$  \\
OC-MAML (ours)  &$85.2$ &$83.5$    &$80.2$ &$84.3$        &$\bf92.5$ &$\bf92.4$     &$\bf91.7$ &$\bf92$\\
\hline

Reptile (w. BN)           &$58.9$ &$56$         &$56.6$ &$62.2$            &$90.4$ &$84.6$        &$88.9$ &$86.7$  \\
FOMAML (w. BN)         &$75.4$ &$72.3$            &$72.2$ &$74.5$           &$91.7$ &$88.6$         &$86$ &$87.8$  \\
MAML (w. BN)            &$83.5$ &$81.3$             &$83.9$ &$77$         &$91.9$ &$90.3$           &$88.7$ &$87.3$  \\
OC-Reptile (w. BN)          &$52$ &$53.2$         &$51.9$ &$51$              &$51.5$ &$51.2$             &$51.1$ &$50.3$  \\
OC-FOMAML (w. BN)      &$74.7$ &$68.5$          &$67$ &$78.7$         &$90.2$ &$85.5$         &$84.3$ &$89.3$  \\
OC-MAML (w. BN) (ours)      &$\bf87.1$ &$\bf86.3$         &$\bf86.8$ &$\bf85.9$      &$92.1$ &$90.7$          &$90.2$ &$91.8$ \\
\hline

\end{tabular}
\end{center}
\end{table*}

\begin{table*}[h]
\begin{center}
\caption{Test accuracies (in $\%$) computed on the class-balanced test sets of the test tasks of the MT-MNIST datasets with $T_{test}=T_{5-8}$. The results are shown for models without BN (top) and with BN (bottom). One-class adaptation sets ($c=0\%$) are used, unless otherwise specified.}
\label{results_meta_mnist_5to8}
\begin{tabular}{| l | l | l | l | l | l | l | l |  l | l | l | l | l |}
\hline
\multicolumn{1}{|c|}{Adaptation set size} &\multicolumn{4}{c|}{$K=2$} &\multicolumn{4}{c|}{$K=10$}\\
\hline
Model $\backslash$ Dataset  &$1$ &$2$ &$3$ &$4$ &$1$ &$2$ &$3$ &$4$  \\
\hline
Reptile         &$60.3$ &$65.4$              &$59.9$ &$57.2$        &$73.6$ &$86.3$                 &$79.1$ &$72.3$  \\
FOMAML          &$76.5$ &$80.2$                &$77.9$ &$72.8$        &$66.2$ &$84.4$                 &$76.9$ &$72.5$  \\
MAML            &$74.8$ &$82.1$             &$73.6$ &$70.7$               &$86.1$ &$93$                    &$90.2$ &$85.7$  \\
OC-Reptile           &$50$ &$50$            &$50$ &$50$                &$50$ &$50$                      &$50$ &$50$  \\
OC-FOMAML       &$54.6$ &$57.3$             &$59$ &$55$            &$53.3$ &$56.7$              &$51$ &$50$  \\
OC-MAML (ours)  &$80.6$ &$\bf91.5$          &$82.1$ &$77.5$                     &$\bf91.5$ &$94.2$     &$91.3$ &$\bf89.8$\\
\hline

Reptile (w. BN)           &$62.3$ &$58.2$         &$60.3$ &$61$            &$85.3$ &$88$        &$88.4$ &$87.2$  \\
FOMAML (w. BN)         &$69.5$ &$75.1$            &$77.3$ &$72.8$           &$86.9$ &$92.3$         &$88.6$ &$85$  \\
MAML (w. BN)            &$84.9$ &$81.8$             &$83.4$ &$76.9$         &$88.6$ &$92.5$           &$90.5$ &$84$  \\
OC-Reptile (w. BN)          &$52$ &$52.2$         &$51.7$ &$53.5$              &$51.1$ &$51.4$             &$50.9$ &$51.8$  \\
OC-FOMAML (w. BN)      &$68.3$ &$85.7$             &$78.5$ &$67.1$               &$83.2$ &$94.7$         &$89.9$ &$82.6$  \\
OC-MAML (w. BN) (ours)      &$\bf85.9$ &$84.8$         &$\bf85.1$ &$\bf82.5$      &$90.5$ &$\bf95.4$          &$\bf91.4$ &$\bf89.8$ \\
\hline

\end{tabular}
\end{center}
\end{table*}

\section{Speeding up OC-MAML} \label{appendix-anil}
A concurrent work \citep{raghu2019rapid} established that MAML's rapid learning of new tasks is dominated by feature reuse. The authors propose the Almost No Inner Loop (ANIL) algorithm, which consists in limiting the task-specific adaptation of MAML (inner loop updates) to the parameters of the model's last layer (the output layer), during meta-training and meta-testing. This leads to a speed up factor of 1.7 over MAML, since ANIL requires the computation of second-order derivative terms only for the last layer's parameters instead of all parameters. ANIL achieves very comparable performance to MAML.

We investigate, whether this simplification of MAML can also speed up OC-MAML, while retaining the same performance. In other words, could we also compute the second-order derivatives, which are required to explicitly optimize for few-shot one-class classification (FS-OCC) (section \ref{theoretical-analysis}), to the last layer's parameters and still reach a model initialization suitable for FS-OCC. Preliminary results of OC-ANIL on the MiniImageNet and Omniglot datasets were very comparable to the results of OC-MAML. Moreover, we conducted the same cosine similarity analysis described in section \ref{results} with ANIL, FOANIL, OC-ANIL and OC-FOANIL and got very consistent results with our findings for the MAML-based algorithms (Table \ref{results_cosine}). This confirms that second-order derivatives have only to be computed for the last layer of the neural network to optimize for FS-OCC, and that OC-ANIL is faster than OC-MAML by a factor of 1.7 \citep{raghu2019rapid} with comparable performance. This modification significantly reduces the computational burden incurred by computing the second-order derivatives for all parameters as done in OC-MAML. Our implementation of OC-ANIL will be published upon paper acceptance.

\end{document}